\documentclass{Interspeech2024}

\usepackage{soul}

\interspeechcameraready

\title{Error Correction by Paying Attention to Both Acoustic and Confidence References for Automatic Speech Recognition
\vspace{-10pt}}

\name[affiliation={1}]{Yuchun}{Shu}
\name[affiliation={2}]{Bo}{Hu}
\name[affiliation={2}]{Yifeng}{He}
\name[affiliation={3}]{Hao}{Shi}
\name[affiliation={1,4}]{Longbiao}{Wang}
\name[affiliation={5}]{Jianwu}{Dang}

\address{
  $^1$Tianjin Key Laboratory of Cognitive Computing and Application, Tianjin University, China \\
  $^2$Baidu, Inc., China 
  $^3$Kyoto University, Japan 
  $^4$Huiyan Technology (Tianjin) Co., Ltd, China \\
  $^5$Shenzhen Institute of Advanced Technolog, Chinese Academy Science, China} 

\email{yuchunshu@tju.edu.cn, longbiao\_wang@tju.edu.cn}

\keywords{speech error correction, speech recognition, confidence estimation}

\begin{document}

\maketitle

\renewcommand{\thefootnote}{\fnsymbol{footnote}}
\footnotetext[1]{Corresponding author: Longbiao Wang.}

\begin{abstract}
Accurately finding the wrong words in the automatic speech recognition (ASR) hypothesis and recovering them well-founded is the goal of speech error correction.
In this paper, we propose a non-autoregressive speech error correction method.
A Confidence Module measures the uncertainty of each word of the N-best ASR hypotheses as the reference to find the wrong word position.
Besides, the acoustic feature from the ASR encoder is also used to provide the correct pronunciation references.
N-best candidates from ASR are aligned using the edit path, to confirm each other and recover some missing character errors.
Furthermore, the cross-attention mechanism fuses the information between error correction references and the ASR hypothesis.
The experimental results show that both the acoustic and confidence references help with error correction.
The proposed system reduces the error rate by 21\% compared with the ASR model.


\end{abstract}

\section{Introduction}

Error correction is a post-processing way to reduce spelling errors in natural language processing (NLP) and automatic speech recognition (ASR). 
The spelling error correction methods for NLP tasks are based on some semantic and grammatical information \cite{zhang2020spelling, liu2021plome, wang2021dynamic, 9689650, liu2022craspell, 10542371}. 
Although NLP error correction methods show advantages, simply applying the NLP spelling correction model to the speech error correction task of ASR may produce adverse effects \cite{ringger1996error, kurata2011training, cucu2013statistical,d2016automatic,tanaka2018neural}. 
Besides, several issues of error correction for ASR have arisen.

The first requirement is that the inference speed needs to meet industrial deployment. 
Although some autoregressive error correction methods \cite{tanaka2018neural, mani2020asr, hu2020deliberation, hu2021transformer, zhu2021improving, liao2023improving} have achieved good error rate reduction, their slow inference speed is unsuitable for online deployment. 
In order to meet the low latency requirements, some non-autoregressive speech error correction works were proposed \cite{leng2021fastcorrect,leng2021fastcorrect2,wang2022towards,leng2023softcorrect}. 
These works are fast in decoding but lack semantic correlation modeling, and the error correction effect is not as good as that of autoregressive. 


Designing the appropriate reference for error correction is also critical.
Some works use the N-best hypothesis of ASR to verify each other and correct errors according to the voting effect \cite{zhu2021improving, leng2021fastcorrect2, leng2023softcorrect, ma2023n}.
Some works use acoustic information or cross-modal information as a reference\cite{hu2020deliberation, hu2021transformer,du2022crossmodal, fan2022acoustic}.
Some works use large-scale text data for pre-training to gain prior knowledge of error correction \cite{dutta2022error, ma2023can}.
These efforts noted reference information for error correction, but were limited in terms of decoding speed or error location detection. 


Accurately finding and modifying the wrong words while avoiding the correct words is the biggest challenge in speech error correction.  
Previous works conduct error detection through two main approaches: implicit detection, where errors are embedded in the correction process \cite{liao2023improving, zhu2021improving}; and explicit detection, where specific errors are identified directly \cite{leng2021fastcorrect,shen2022mask,du2022crossmodal,yeen23_interspeech}.
Implicit error detection offers the flexibility of model learning but lacks clear feedback for training, while explicit error detection provides clear training signals but may introduce new errors if inaccurate. 
Therefore, a high-accuracy error detection module is necessary. 

In this paper, we propose a speech error correction model to alleviate the above-mentioned issues within a model. 
The proposed method is designed as non-autoregressive to reduce the inference latency. 
The ASR encoder encodes the input speech features. 
The deep encoder layers contain refined acoustic and semantic information, which can provide the correct pronunciation reference and is suitable for non-autoregressive correction. 
We also propose to use a Confidence Module to measure the uncertainty of each input word embedding. 
The estimated confidence score is used as the reference for error correction to provide information on wrong word positions. 
Furthermore, the N-best ASR hypotheses are aligned during correction, to recover some deletion errors and achieve variable-length error correction. 
The cross-attention mechanism is used to fuse the information between references and the ASR hypothesis embedding. 
Finally, the embeddings fused with different references are added and reclassified to get the correction results.

\section{Preliminaries}
\subsection{Acoustic Feature} 
\vspace{-5pt}
Attention-based encoder-decoder end-to-end ASR models directly convert audio to text. 
The encoder and decoder mainly act as acoustic and language models, respectively. 
The encoder takes the speech features as input, extracting acoustic features layer by layer, and outputs the acoustic hidden variables \cite{shi2024investigation}. 
The decoder, acting as a language model, converts the input acoustic hidden variables into the output text sequence. 
There are information differences between the deep and shallow features of the encoder. 
Shallower layers contain richer acoustic information, such as speech style, tone, and timbre. 
In comparison, deeper layers encode more refined pronunciation details, which aligns with the information required for ASR tasks.

\begin{figure}[ht]
  \centering
  \centering{\includegraphics[width=1.\linewidth]
  {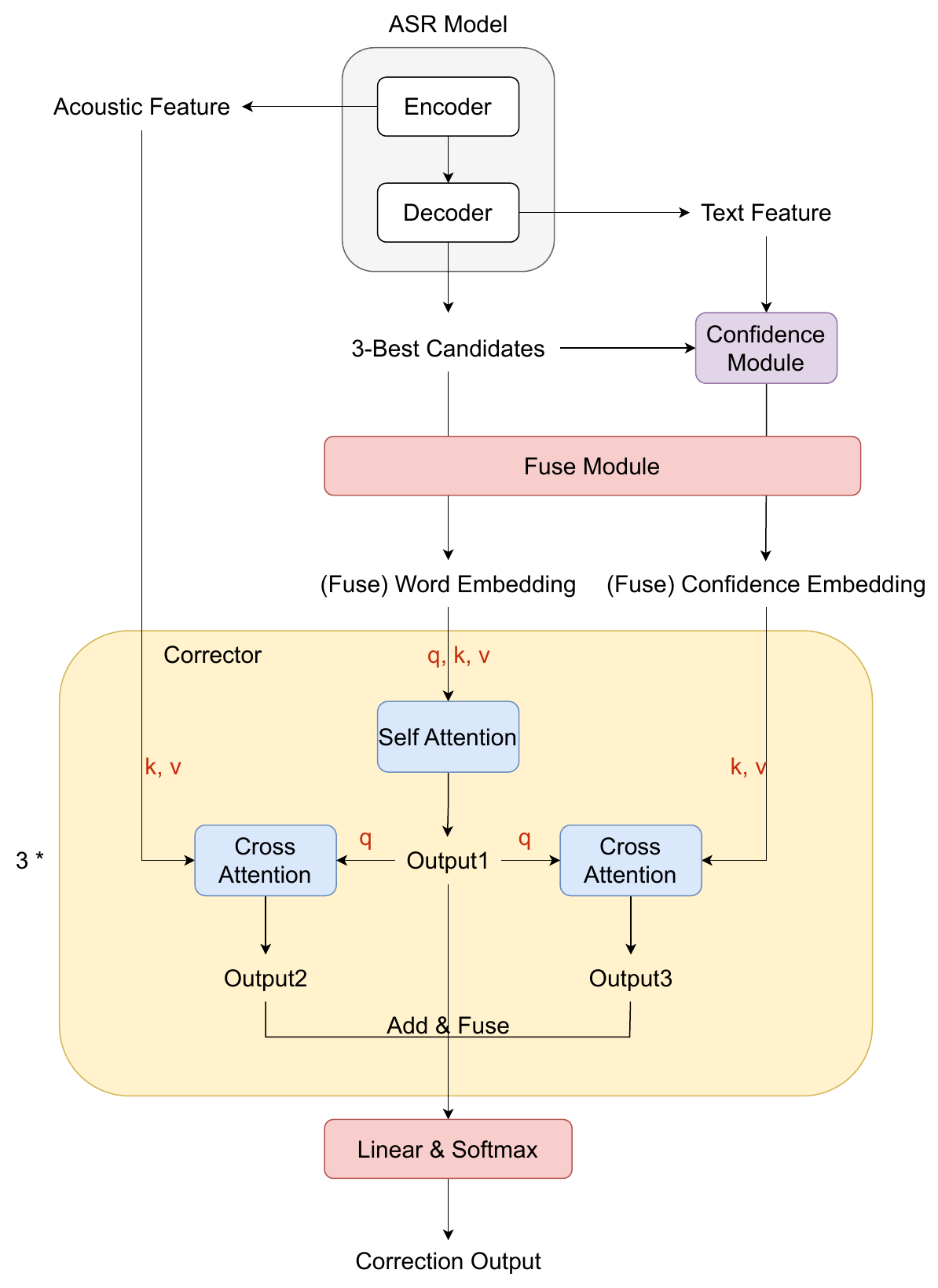}}
  \vspace{-20pt}
  \caption{Flowchart of proposed speech error correction model.}
  \vspace{-15pt}
  \label{correct}
\end{figure}

\begin{figure}[ht]
  \centering
  \centering{\includegraphics[width=0.9\linewidth]
  {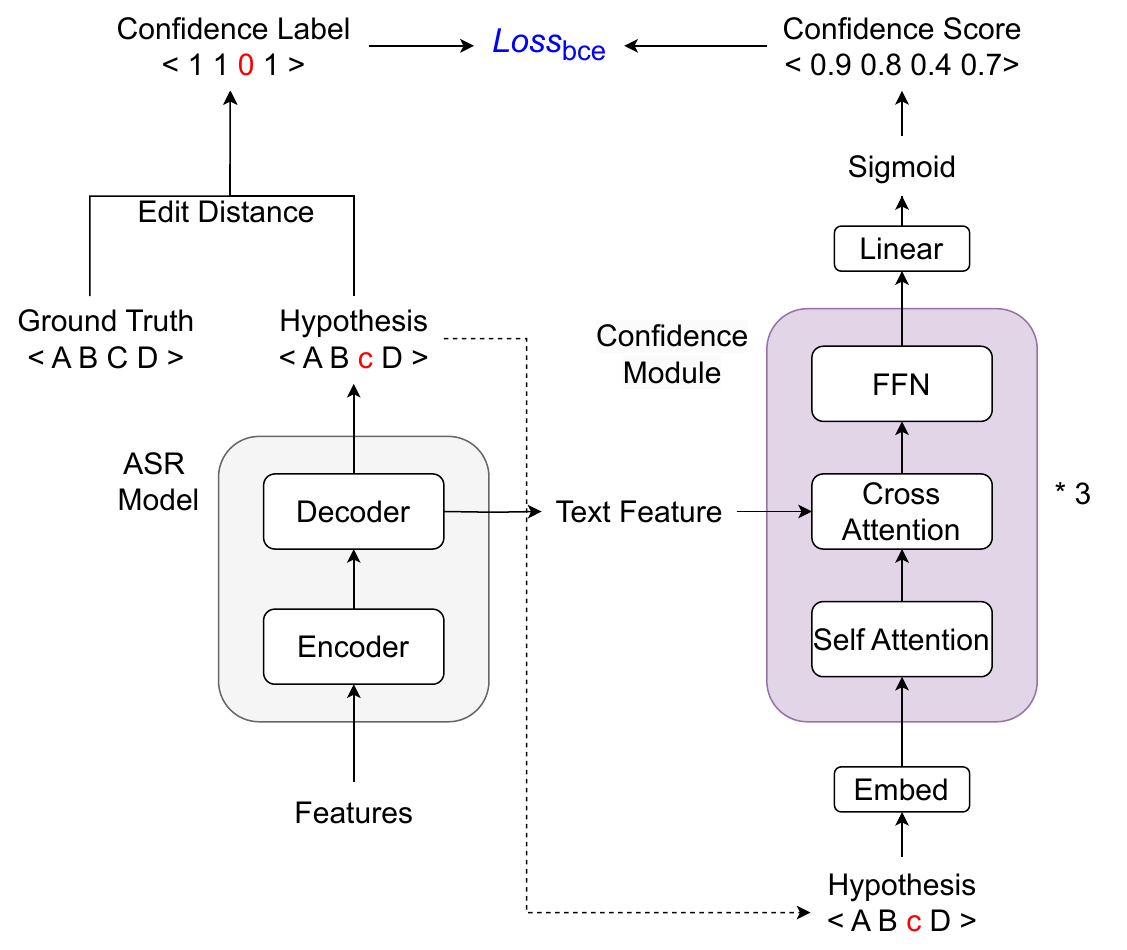}}
  \vspace{-10pt}
  \caption{Training of Confidence Module.}
  \vspace{-15pt}
  \label{zhixin}
\end{figure}

\subsection{Confidence Estimation}
\vspace{-5pt}
Confidence estimation is important in many fields, including ASR\cite{9056075,qiu2021learning,shi2022multi,shi2023accurate}. 
In recent years, although the accuracy of end-to-end ASR has improved, the quality of confidence estimation has collapsed. 
In terms of confidence accuracy, the posterior probability given by the Softmax layer after the speech recognition decoder proves to be sharp and of poor quality\cite{li2021confidence}. 
In other words, when the ASR model gets the wrong decoding result, the model is still confident in its answer and has a high posterior probability pointing to the current answer.

\section{Proposed Method}
The confidence score often plays the role of uncertainty measurement. 
To solve the problem of overconfidence in ASR models, we add the confidence estimation task to the speech error correction system to provide the correct probability of each predicted token as the reference information. 
Besides, most previous speech error correction works ignore the speaker's acoustic information, which provides frame-level discriminative information.
Thus, we incorporate acoustic information from the ASR model to provide the pronunciation reference for wrong words.

\subsection{Correction with Both Acoustic and Confidence References}
\vspace{-5pt}
The overall flowchart of the error correction model proposed in this work is shown in Figure~\ref{correct}. 
The error correction adopts both the acoustic and confidence references to correct the ASR hypothesis. 
Besides, the error correction adopts the 3-best candidates from ASR as inputs to obtain better error correction accuracy.

\noindent\textbf{Acoustic reference}: 
When extracting acoustic features, the last layer output vector of the encoder is too close to the classification task of the decoder. 
Thus, the error correction result will be biased to the recognition result. 
Conversely, the output vector of very shallow layers contains too much irrelevant information. 
In the trade-off between the shallow and deep layers, the acoustic features from the intermediate layer (10th out of 12 encoder layers) are used to provide more refined acoustic information while different from the ASR task.

\noindent\textbf{Confidence reference}: 
A Confidence Module (CEM) is adopted as the error detection module, which inputs the recognition result embedding and text feature to provide the confident representation of each word. 
After being aligned by the editing distance, the three candidates enter the Confidence Module to evaluate the confidence score. 
Three candidates take the hidden variable of the last layer of CEM respectively as the reference for error correction. 
A linear layer is used as the interpolation coefficient fusion module, and three weights are learned according to the word embedding and confidence embedding of ASR 3-best results, after concatenation as input.    
The three candidate word/confidence embeddings are weighted summed by the three weights to obtain the fused word/confidence embeddings, which are input into the error correction module. 

The error correction module adopts the non-autoregressive method for fast decoding speed. 
The error correction module is a three-layer cross-attention decoder structure. 
The fused word embedding is first calculated by self-attention, and then two cross-attention mechanisms are performed with acoustic feature and fused confidence embedding, respectively. 
In cross-attention calculations, Query is derived from the word embedding, and Key and Value are derived from acoustic feature or fused confidence embedding. 
The outputs of the three attention calculations are added and then reclassified to get the error correction result.

\subsection{Confidence Module (CEM)}
\vspace{-5pt}
An attention-based CEM is trained to provide confidence scores for recognition results. 
The structure and training flow of the CEM are illustrated in Figure \ref{zhixin}. 
The CEM consists of a three-layer decoder (the same as the Transformer) and a Sigmoid function that predicts the confidence score of the ASR hypothesis. 
The input of CEM is the embedding extracted from the ASR hypothesis sequence and the text feature output from the middle layer (5th out of 6) of the ASR decoder. 
The edit distance is calculated between each hypothesis sequence relative to the ground truth sequence. 
The correct token is specified as 1, while the replacement or inserted token is specified as 0, making the alignment of the edit distance calculation available as the target for CEM training. 
\begin{figure}[h]
  \centering
  \centering{\includegraphics[width=0.75\linewidth]
  {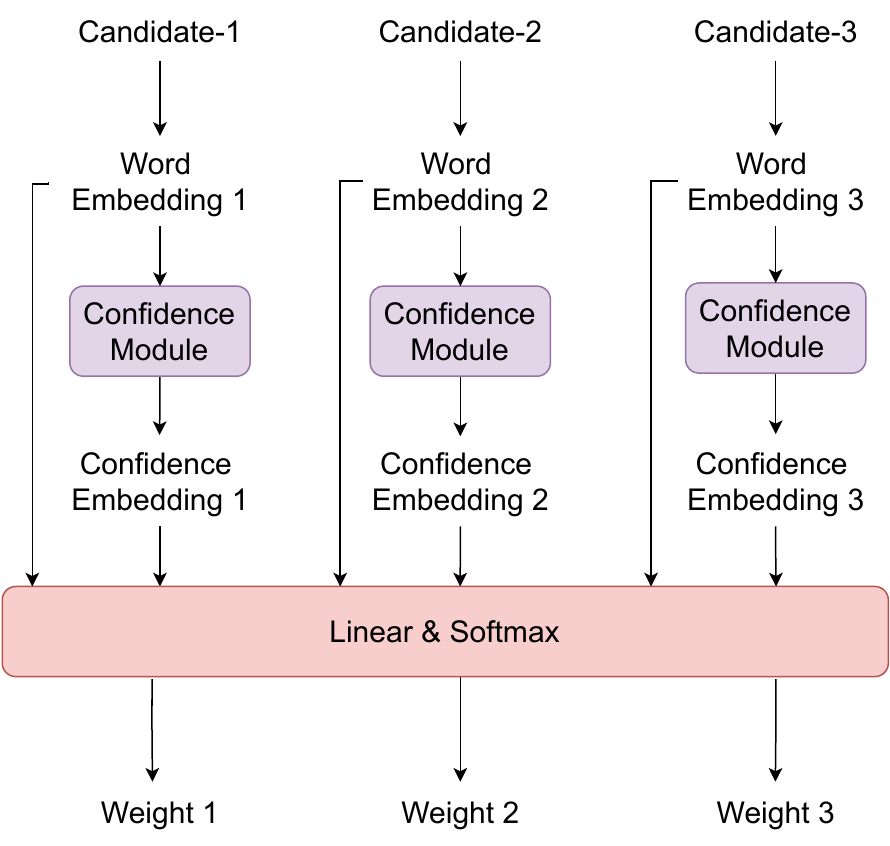}}
  \vspace{-10pt}
  \caption{Flowchart of the N-best Fusion Module.}
  \vspace{-15pt}
  \label{fuse}
\end{figure}
For example, if the ground truth sequence is $<A B C D>$ and the ASR hypothesis is $<A B c D>$, then the binary target sequence is $c = <1, 1, 0, 1>$. 
The forward process is in full sequence, not autoregressive mode. 
For each N-best hypothesis, the CEM is trained to minimize the binary cross-entropy loss as follows: 
\begin{equation}
    \footnotesize
    \setlength{\abovedisplayskip}{2pt}
    \setlength{\belowdisplayskip}{2pt}
	L(c, p) = -\frac{1}{L}\sum_{l=1}^{L}{(c_l \times log(p_l) + ( 1 - c_l) \times log(1 - p_l))}
	\label{eqloss}
\end{equation}
where $p$ is the estimated confidence score, $c$ is the target sequence, $L$ is the length of the hypothesis. 

\noindent\textbf{CEM pre-training}: The CEM carries out pre-training before constructing the speech error correction module. The hidden variables of the last layer of CEM are extracted as confidence features and input into the speech error correction model.
During the training of the speech error correction model, the parameters of CEM are frozen and do not participate in the training and parameter updating.

\subsection{N-best Fusion Module}
\vspace{-5pt}
As shown in Figure \ref{fuse}, after obtaining the N-best hypotheses from ASR ($N = 3$ in this experiment), all candidates were aligned according to the method proposed by FastCorrect2 \cite{leng2021fastcorrect2}. 
After alignment, candidates of the same length are then passed over the Confidence Module respectively to obtain confidence embedding. According to the word embedding ($\text{emb}^{w}$) and confidence embedding ($\text{emb}^{c}$) of N-best candidates, a linear layer is used to learn the weight of each candidate (${w}_{i}$). 
Finally, the fusion word embedding and fusion confidence embedding are weighted summed by the learned weight of each candidate. 
Multiple candidates are fused before entering the error correction module, avoiding additional attention computing time and further reducing error rates. 
Adding N-best information for ASR also supports the recovery of deleted errors to a certain extent and achieves variable-length error correction. 
The fusion is as follows: 
\begin{equation}
    \setlength{\abovedisplayskip}{2pt}
    \setlength{\belowdisplayskip}{2pt}
    \begin{split}
        &\text{$\text{emb}^{w}_{f}$}=\sum\nolimits_{i=1}^{N} {w}_{i} \times \text{$\text{emb}^{w}_{i}$}\\
        &\text{$\text{emb}^{c}_{f}$}=\sum\nolimits_{i=1}^{N}{w}_{i} \times \text{$\text{emb}^{c}_{i}$}
    \end{split}
	\label{eq4_1}
\end{equation}
where $\text{$\text{emb}^{w}_{f}$}$ and $\text{$\text{emb}^{c}_{f}$}$ are the fused embedding of word embedding and confidence embedding.

\section{Experimental Setup}

\subsection{Experimental Settings for ASR}
\vspace{-5pt}
We evaluate our system on the AISHELL-1 \cite{bu2017aishell} Mandarin ASR dataset (150/10/5 hours for train/dev/test). 
The ASR model employs a Conformer architecture with SpecAugment, speed perturbation, and a joint decoding language model. 
The hyper-parameters of the ASR model follow the Wenet codebase \cite{yao2021wenet} (12-encoder/6-decoder layers, 256d 4-head attention, 1024d feed-forward network). 
The vocabulary size was 18532.


\subsection{Experimental Settings for Error Correction}
\vspace{-5pt}
This section compares four baselines of error correction, two autoregressive and two non-autoregressive methods. 
The autoregressive methods are both standard error correction methods based on Transformer, one using only the best candidate (B1) \cite{vaswani2017attention}, and another using the aligned N-best candidates as input (B2) \cite{zhu2021improving}. 
Two non-autoregressive error correction methods from Microsoft (FastCorrect (B3) \cite{leng2021fastcorrect}, FastCorrect2 (B4) \cite{leng2021fastcorrect2}) that calculate edit distance and utilize a length predictor. 
FastCorrect uses only the best candidates, and FastCorrect2 incorporates aligned N-best candidates. 
These baselines' error correction experimental results are from \cite{leng2023softcorrect}. 
All baseline error correction models were pre-trained with a 400M unpaired text dataset and then fine-tuned with the AISHELL-1 dataset.

Our proposed method requires acoustic features and confidence scores, necessitating a 1,000-hour speech dataset for pre-training. 
Due to these pre-training data differences, error correction comparisons should be considered for reference only. 
During training, the CEM is first pre-trained according to section 3.2. Then the CEM parameters are frozen and other modules of the error correction model are trained (same as ASR, first pre-trained with the 1,000-hour speech dataset then fine-tuned with AISHELL-1).

Both the confidence estimation and error correction modules utilize a 3-layer cross-attention decoder with a model dimension of 384, 768-dimensional feed-forward networks, and 6-head attention. 
The error correction module performs the cross-attention calculation twice for each layer.

\subsection{Evaluation Metrics}
\vspace{-5pt}
Beyond standard classification metrics (accuracy, precision, recall, specificity, and f1-score), Normalized Cross-Entropy (NCE) was used to gauge the quality of confidence scores \cite{siu1997improved}. 
If confidence scores for all tokens are gathered as $p = [p_1,...,p_N]$ where $p_n \in [0,1]$, and the corresponding target confidence is set to $c = [c_1,...,c_N]$ where $c_n \in \{0,1\}$, the NCE is computed as follows: 
\begin{equation}
    \setlength{\abovedisplayskip}{2pt}
    \setlength{\belowdisplayskip}{2pt}
	NCE(c, p) = \frac{H(c) - H(c, p)}{H(c)}
	\label{NCE}
\end{equation}
where $H(c)$ is the entropy of the target sequence and $H(c, p)$ is the binary cross-entropy between the target and the estimated confidence scores. 
NCE metric measures how closely the estimated confidence score reflects the true probability of a recognized word being correct. 
When confidence estimation is systematically better than the word correct ratio of ASR, NCE is positive. 
The closer the NCE is to 1, the higher the quality of the confidence estimation.

The speech error correction method is evaluated using the following metrics: character error rate (\textbf{CER}) after correction, character error rate reduction (\textbf{CERR}) relative to the ASR hypothesis, and inference latency (\textbf{Latency}) measured on one NVIDIA V100 GPU.

\begin{table}[h]   
	\begin{center} 
		\centering
		\caption{Confidence estimation result, the threshold for binary classification is set to 0.5.}
        \vspace{-10pt}
		\label{table_correct}
		\setlength{\tabcolsep}{2pt}
		\begin{tabular}{m{134pt}|m{40pt}<{\centering}m{40pt}<{\centering}}
			\toprule[1.5pt]   
			\textbf{Confidence } & \multicolumn{2}{c}
{\textbf{AISHELL}}  \\
			\textbf{Estimation} & \textbf{Test} & \textbf{Dev} \\
			\midrule[1pt] 
            Accuracy of ASR & 0.951 & 0.958 \\
            \midrule[1pt]
			Accuracy  & 0.964 & 0.968 \\ 
            Precision & 0.973 & 0.976 \\
            Recall & 0.990 & 0.992 \\
            Specifity & 0.485 & 0.456\\
            F1-score & 0.981 & 0.984 \\
            NCE & 0.500 & 0.494 \\
			\bottomrule[1.5pt]
		\end{tabular} 
	\end{center}   
 \vspace{-15pt}
\end{table}

\begin{table}[h]   
	\begin{center} 
		\centering
		\caption{Experimental results of different error correction models in AISHELL test set.}
  \vspace{-10pt}
		\label{table_aishell}
		\setlength{\tabcolsep}{2pt}
  \resizebox{\linewidth}{!}{
		\begin{tabular}{m{75pt}|m{23pt}<{\centering}m{25pt}<{\centering}m{23pt}<{\centering}m{25pt}<{\centering}m{28pt}<{\centering}}
			\toprule[1.5pt]   
			\textbf{Model} & \multicolumn{5}{c}
{\textbf{AISHELL}}  \\
			~ & \multicolumn{2}{c}{\textbf{Test}} & \multicolumn{2}{c}{\textbf{Dev}} & \textbf{Latency}\\
			~ & \textbf{CER} & \textbf{CERR} & \textbf{CER}	 &\textbf{CERR} \\
            ~ & \textbf{(\%)} & \textbf{(\%)} & \textbf{(\%)}	 &\textbf{(\%)} & (ms/sent)\\
			\midrule[1pt]
			ASR No Correction  & 4.83 & - & 4.46 & - & -\\  
            B1: AR Correct & 4.07 & 15.73 & 3.79 & 15.02 & 119.0\\
            B2: AR N-Best & 3.94 & 18.43 & 3.68 & 17.49 & 121.6\\

            B3: FastCorrect  & 4.16 & 13.87 & 3.89 & 12.78 & 16.2\\ 
            B4: FastCorrect2  & 4.11 & 14.91 & 3.78 & 15.25 & 23.1\\ 
			\midrule[1pt] 
			ASR No Correction  & 4.91 & - & 4.21 & - & - \\ 
			M1: 1-best + CEM & 4.37 & 11.00 & 3.85 & 8.55 & -\\
			M2: 1-best + Acoustic  & 4.12 & 16.09 & 3.66 & 13.06 & -\\
			M3: 1-best + CEM + Acoustic  & 3.94 & 19.80 & 3.45 & 18.05 & -\\ 
			M4: 3-best + CEM + Acoustic & 3.88$\star$ & 21.00 & 3.40$\triangleleft$ & 19.20 & 25.0 \\
			\bottomrule[1.5pt]
            \multicolumn{6}{r}{($\star$: p-value $<$ 0.05 against B4; $\triangleleft$: p-value $<$ 0.01 against B4)}
		\end{tabular} 
      }
	\end{center} 
 \vspace{-20pt}
\end{table}

\section{Experimental Results}
Table \ref{table_correct} shows the confidence estimation result. 
It demonstrates the high accuracy and F1-score of the Confidence Module for binary classification. 
The F1-score is usually between 0.5 and 0.6 for error detection modules of four baselines \cite{leng2023softcorrect}. 
For confidence estimation, NCE between 0.2 and 0.4 can indicate the effectiveness of the module \cite{li2021confidence}. 
The proposed CEM surpassed most confidence estimation approaches, boasting an NCE value exceeding 0.49 and an F1-score above 0.98. 
This suggests that the attention-based CEM effectively distinguishes between correct and incorrect tokens, providing valuable information for error correction.

Table \ref{table_aishell} summarizes the performance of various error correction systems. 
Our proposed methods (M1-M4) outperform the baselines (B1-B4) regarding CERR. 
M1, which leverages confidence information alongside the ASR 1-best hypothesis, achieves a CERR of 11.00\% and 8.55\% on the test and dev sets, respectively. 
Confidence scores guide the error correction module to prioritize fixing likely mistakes in the speech recognition output. 
Acoustic information (M2) proves even more beneficial, which offers the pronunciation reference for wrong words, achieving a CERR of 16.09\% and 13.06\%. 
\begin{figure}[ht]
  \centering
  \centering{\includegraphics[width=\linewidth]
  {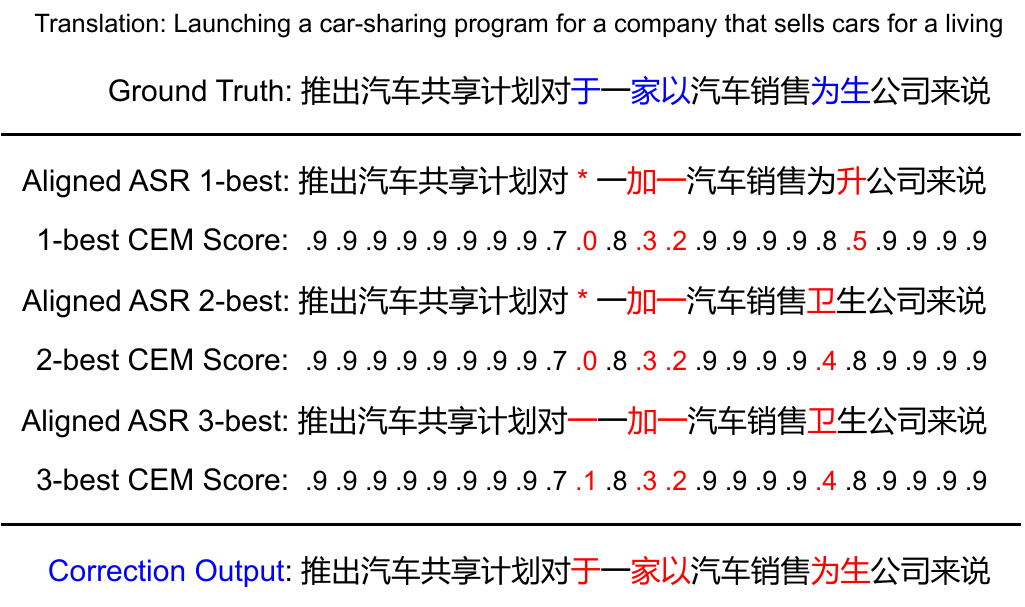}}
  \vspace{-20pt}
  \caption{An example of the error correction process of the ASR hypothesis.}
  \vspace{-15pt}
  \label{case}
\end{figure}
Combining both acoustic feature and confidence information (M3) leads to further gains (19.80\% and 18.05\% CERR), suggesting the complementary between different references. 
Finally, incorporating the 3-best ASR hypotheses (M4) yields an additional 1.2\% improvement (21.00\% and 19.20\% CERR), highlighting the effectiveness of N-best voting for error correction.


Non-autoregressive models are significantly faster for real-world use (4--10$\times$) than autoregressive models, but they trade some accuracy for this speed.  This is because they can't consider previously decoded information during character prediction, making it harder to capture the semantic relationships of the sentence.
According to the inference latency in Table \ref{table_aishell}, our proposed error correction system maintains the speed advantage in a non-autoregressive manner and surpasses the error correction performance of autoregressive baselines (B1, B2). 
Although the latency increases slightly (acceptable in industrial deployments) compared to non-autoregressive baselines (B3, B4), there is a statistically significant reduction in CER.

An example of the error correction is given in Figure \ref{case}. 
Both the 1-best and 2-best candidates have one deletion and three substitution errors. 
For the 3-best candidate, there are four substitution errors. 
By aligning all three hypotheses, the 1-best and 2-best candidates complement a blank token (*) at the location of the deletion error. 
The CEM provides confidence scores for each hypothesis, pinpointing potential error locations.
For accurate words, the confidence scores were above 0.9. 
For uncertain words, the scores were below 0.5. 
Placeholder blank ID had a confidence score of only 0.02. 
Referring to the acoustic feature, confidence information, and the voting effect of three candidates, the error correction module effectively recovers the error token in the ASR hypothesis. 

This work adopts the edit distance alignment method, which can potentially recover part of the deletion error by aligning the 3-best candidates. 
However, if all three candidates miss the same character, the system cannot perform variable-length recovery, and there are still some limitations.

\section{Conclusion and Future Works}
In this paper, we proposed a new non-autoregressive speech error correction method for automatic speech recognition. 
The acoustic feature from the ASR encoder and the confidence information of each word are used as references to assist the error correction model in determining the error location and correct pronunciation. 
Furthermore, the proposed method further improved the performance by using the N-best hypotheses from ASR to confirm each other. 
Experimental results show that the proposed system can reduce the error rate by 21\% compared with the ASR model, and the error correction effect is remarkable. 
Future work will further explore variable-length error correction solutions for identified deletion errors.

\section{Acknowledgments}
This work was supported by the National Natural Science Foundation of China under Grant 62176182 and U23B2053.

\bibliographystyle{IEEEtran}
\bibliography{reference}

\begin{thebibliography}{10}
\providecommand{\url}[1]{#1}
\csname url@samestyle\endcsname
\providecommand{\newblock}{\relax}
\providecommand{\bibinfo}[2]{#2}
\providecommand{\BIBentrySTDinterwordspacing}{\spaceskip=0pt\relax}
\providecommand{\BIBentryALTinterwordstretchfactor}{4}
\providecommand{\BIBentryALTinterwordspacing}{\spaceskip=\fontdimen2\font plus
\BIBentryALTinterwordstretchfactor\fontdimen3\font minus
  \fontdimen4\font\relax}
\providecommand{\BIBforeignlanguage}[2]{{%
\expandafter\ifx\csname l@#1\endcsname\relax
\typeout{** WARNING: IEEEtran.bst: No hyphenation pattern has been}%
\typeout{** loaded for the language `#1'. Using the pattern for}%
\typeout{** the default language instead.}%
\else
\language=\csname l@#1\endcsname
\fi
#2}}
\providecommand{\BIBdecl}{\relax}
\BIBdecl

\bibitem{zhang2020spelling}
S.~Zhang, H.~Huang, J.~Liu, and H.~Li, ``{Spelling Error Correction with
  Soft-Masked BERT},'' in \emph{Proceedings of the 58th Annual Meeting of the
  Association for Computational Linguistics}, 2020, pp. 882--890.

\bibitem{liu2021plome}
S.~Liu, T.~Yang, T.~Yue, F.~Zhang, and D.~Wang, ``{PLOME: Pre-training with
  misspelled knowledge for Chinese spelling correction},'' in \emph{Proceedings
  of the 59th Annual Meeting of the Association for Computational Linguistics
  and the 11th International Joint Conference on Natural Language Processing
  (Volume 1: Long Papers)}, 2021, pp. 2991--3000.

\bibitem{wang2021dynamic}
B.~Wang, W.~Che, D.~Wu, S.~Wang, G.~Hu, and T.~Liu, ``{Dynamic connected
  networks for Chinese spelling check},'' in \emph{Findings of the Association
  for Computational Linguistics: ACL-IJCNLP 2021}, 2021, pp. 2437--2446.

\bibitem{9689650}
H.~Shi, L.~Wang, S.~Li, C.~Fan, J.~Dang, and T.~Kawahara, ``Spectrograms
  fusion-based end-to-end robust automatic speech recognition,'' in \emph{Proc.
  APSIPA ASC}, 2021, pp. 438--442.

\bibitem{liu2022craspell}
S.~Liu, S.~Song, T.~Yue, T.~Yang, H.~Cai, T.~Yu, and S.~Sun, ``{CRASpell: A
  contextual typo robust approach to improve Chinese spelling correction},'' in
  \emph{Proc. ACL}, 2022, pp. 3008--3018.

\bibitem{10542371}
H.~Shi, M.~Mimura, and T.~Kawahara, ``Waveform-domain speech enhancement using
  spectrogram encoding for robust speech recognition,'' \emph{IEEE/ACM
  Transactions on Audio, Speech, and Language Processing}, vol.~32, pp.
  3049--3060, 2024.

\bibitem{ringger1996error}
E.~K. Ringger and J.~F. Allen, ``Error correction via a post-processor for
  continuous speech recognition,'' in \emph{Proc. ICASSP}, vol.~1.\hskip 1em
  plus 0.5em minus 0.4em\relax IEEE, 1996, pp. 427--430.

\bibitem{kurata2011training}
G.~Kurata, N.~Itoh, and M.~Nishimura, ``Training of error-corrective model for
  asr without using audio data,'' in \emph{Proc. ICASSP}.\hskip 1em plus 0.5em
  minus 0.4em\relax IEEE, 2011, pp. 5576--5579.

\bibitem{cucu2013statistical}
H.~Cucu, A.~Buzo, L.~Besacier, and C.~Burileanu, ``{Statistical error
  correction methods for domain-specific ASR systems},'' in \emph{Statistical
  Language and Speech Processing: First International Conference, SLSP 2013,
  Tarragona, Spain, July 29-31, 2013. Proceedings 1}.\hskip 1em plus 0.5em
  minus 0.4em\relax Springer, 2013, pp. 83--92.

\bibitem{d2016automatic}
L.~F. D’Haro and R.~E. Banchs, ``Automatic correction of asr outputs by using
  machine translation,'' in \emph{Proc. Interspeech}, vol. 2016, 2016, pp.
  3469--3473.

\bibitem{tanaka2018neural}
T.~Tanaka, R.~Masumura, H.~Masataki, and Y.~Aono, ``{Neural Error Corrective
  Language Models for Automatic Speech Recognition.}'' in \emph{Proc.
  INTERSPEECH}, 2018, pp. 401--405.

\bibitem{mani2020asr}
A.~Mani, S.~Palaskar, N.~V. Meripo, S.~Konam, and F.~Metze, ``{ASR error
  correction and domain adaptation using machine translation},'' in \emph{Proc.
  ICASSP}.\hskip 1em plus 0.5em minus 0.4em\relax IEEE, 2020, pp. 6344--6348.

\bibitem{hu2020deliberation}
K.~Hu, T.~N. Sainath, R.~Pang, and R.~Prabhavalkar, ``Deliberation model based
  two-pass end-to-end speech recognition,'' in \emph{Proc. ICASSP}.\hskip 1em
  plus 0.5em minus 0.4em\relax IEEE, 2020, pp. 7799--7803.

\bibitem{hu2021transformer}
K.~Hu, R.~Pang, T.~N. Sainath, and T.~Strohman, ``Transformer based
  deliberation for two-pass speech recognition,'' in \emph{Proc. SLT}.\hskip
  1em plus 0.5em minus 0.4em\relax IEEE, 2021, pp. 68--74.

\bibitem{zhu2021improving}
L.~Zhu, W.~Liu, L.~Liu, and E.~Lin, ``Improving asr error correction using
  n-best hypotheses,'' in \emph{Proc. ASRU}.\hskip 1em plus 0.5em minus
  0.4em\relax IEEE, 2021, pp. 83--89.

\bibitem{liao2023improving}
J.~Liao, S.~Eskimez, L.~Lu, Y.~Shi, M.~Gong, L.~Shou, H.~Qu, and M.~Zeng,
  ``Improving readability for automatic speech recognition transcription,''
  \emph{ACM Transactions on Asian and Low-Resource Language Information
  Processing}, vol.~22, no.~5, pp. 1--23, 2023.

\bibitem{leng2021fastcorrect}
Y.~Leng, X.~Tan, L.~Zhu, J.~Xu, R.~Luo, L.~Liu, T.~Qin, X.~Li, E.~Lin, and
  T.-Y. Liu, ``{Fastcorrect: Fast error correction with edit alignment for
  automatic speech recognition},'' \emph{Proc. NIPS}, vol.~34, pp.
  21\,708--21\,719, 2021.

\bibitem{leng2021fastcorrect2}
Y.~Leng, X.~Tan, R.~Wang, L.~Zhu, J.~Xu, W.~Liu, L.~Liu, X.-Y. Li, T.~Qin,
  E.~Lin \emph{et~al.}, ``{FastCorrect 2: Fast Error Correction on Multiple
  Candidates for Automatic Speech Recognition},'' in \emph{Proc. EMNLP}, 2021,
  pp. 4328--4337.

\bibitem{wang2022towards}
X.~Wang, Y.~Liu, J.~Li, V.~Miljanic, S.~Zhao, and H.~Khalil, ``Towards
  contextual spelling correction for customization of end-to-end speech
  recognition systems,'' \emph{IEEE/ACM TASLP}, vol.~30, pp. 3089--3097, 2022.

\bibitem{leng2023softcorrect}
Y.~Leng, X.~Tan, W.~Liu, K.~Song, R.~Wang, X.-Y. Li, T.~Qin, E.~Lin, and T.-Y.
  Liu, ``{Softcorrect: Error correction with soft detection for automatic
  speech recognition},'' in \emph{Proc. AAAI}, vol.~37, no.~11, 2023, pp.
  13\,034--13\,042.

\bibitem{ma2023n}
R.~Ma, M.~J. Gales, K.~M. Knill, and M.~Qian, ``N-best t5: Robust asr error
  correction using multiple input hypotheses and constrained decoding space,''
  \emph{arXiv preprint arXiv:2303.00456}, 2023.

\bibitem{du2022crossmodal}
J.~Du, S.~Pu, Q.~Dong, C.~Jin, X.~Qi, D.~Gu, R.~Wu, and H.~Zhou, ``{Cross-Modal
  ASR Post-Processing System for Error Correction and Utterance Rejection},''
  2022.

\bibitem{fan2022acoustic}
R.~Fan, G.~Ye, Y.~Gaur, and J.~Li, ``Acoustic-aware non-autoregressive spell
  correction with mask sample decoding,'' \emph{arXiv preprint
  arXiv:2210.08665}, 2022.

\bibitem{dutta2022error}
S.~Dutta, S.~Jain, A.~Maheshwari, S.~Pal, G.~Ramakrishnan, and P.~Jyothi,
  ``Error correction in asr using sequence-to-sequence models,'' \emph{arXiv
  preprint arXiv:2202.01157}, 2022.

\bibitem{ma2023can}
R.~Ma, M.~Qian, P.~Manakul, M.~Gales, and K.~Knill, ``Can generative large
  language models perform asr error correction?'' \emph{arXiv preprint
  arXiv:2307.04172}, 2023.

\bibitem{shen2022mask}
K.~Shen, Y.~Leng, X.~Tan, S.~Tang, Y.~Zhang, W.~Liu, and E.~Lin, ``Mask the
  correct tokens: An embarrassingly simple approach for error correction,'' in
  \emph{Proceedings of the 2022 Conference on Empirical Methods in Natural
  Language Processing}, 2022, pp. 10\,367--10\,380.

\bibitem{yeen23_interspeech}
H.-Y. Yeen, M.-J. Kim, and M.-W. Koo, ``I learned error, i can fix it! : A
  detector-corrector structure for asr error calibration,'' in \emph{Proc.
  INTERSPEECH}, 2023, pp. 2693--2697.

\bibitem{shi2024investigation}
H.~Shi and T.~Kawahara, ``Exploration of adapter for noise robust automatic
  speech recognition,'' \emph{arXiv preprint arXiv:2402.18275}, 2024.

\bibitem{9056075}
A.~K. Punnoose, ``Substate detection based confidence scoring in speech
  recognition,'' in \emph{2020 National Conference on Communications (NCC)},
  2020, pp. 1--5.

\bibitem{qiu2021learning}
D.~Qiu, Q.~Li, Y.~He, Y.~Zhang, B.~Li, L.~Cao, R.~Prabhavalkar, D.~Bhatia,
  W.~Li, K.~Hu \emph{et~al.}, ``Learning word-level confidence for subword
  end-to-end asr,'' in \emph{Proc. ICASSP}.\hskip 1em plus 0.5em minus
  0.4em\relax IEEE, 2021, pp. 6393--6397.

\bibitem{shi2022multi}
L.~Shi, Z.~Wang, Y.~Du, Y.~Wang, Y.~Chen, J.~Zhao, and K.~Wang, ``A
  multi-feature fusion based confidence estimation model for chinese end-to-end
  speech recognition,'' in \emph{Proc. ICCSCT)}, vol. 12506.\hskip 1em plus
  0.5em minus 0.4em\relax SPIE, 2022, pp. 535--541.

\bibitem{shi2023accurate}
X.~Shi, H.~Luo, Z.~Gao, S.~Zhang, and Z.~Yan, ``Accurate and reliable
  confidence estimation based on non-autoregressive end-to-end speech
  recognition system,'' \emph{arXiv preprint arXiv:2305.10680}, 2023.

\bibitem{li2021confidence}
Q.~Li, D.~Qiu, Y.~Zhang, B.~Li, Y.~He, P.~C. Woodland, L.~Cao, and T.~Strohman,
  ``Confidence estimation for attention-based sequence-to-sequence models for
  speech recognition,'' in \emph{Proc. ICASSP}.\hskip 1em plus 0.5em minus
  0.4em\relax IEEE, 2021, pp. 6388--6392.

\bibitem{bu2017aishell}
H.~Bu, J.~Du, X.~Na, B.~Wu, and H.~Zheng, ``{Aishell-1: An open-source mandarin
  speech corpus and a speech recognition baseline},'' in \emph{Proc.
  O-COCOSDA}.\hskip 1em plus 0.5em minus 0.4em\relax IEEE, 2017, pp. 1--5.

\bibitem{yao2021wenet}
Z.~Yao, D.~Wu, X.~Wang, B.~Zhang, F.~Yu, C.~Yang, Z.~Peng, X.~Chen, L.~Xie, and
  X.~Lei, ``{Wenet: Production oriented streaming and non-streaming end-to-end
  speech recognition toolkit},'' \emph{arXiv preprint arXiv:2102.01547}, 2021.

\bibitem{vaswani2017attention}
A.~Vaswani, N.~Shazeer, N.~Parmar, J.~Uszkoreit, L.~Jones, A.~N. Gomez,
  {\L}.~Kaiser, and I.~Polosukhin, ``Attention is all you need,''
  \emph{Advances in neural information processing systems}, vol.~30, 2017.

\bibitem{siu1997improved}
M.-h. Siu, H.~Gish, and F.~Richardson, ``Improved estimation, evaluation and
  applications of confidence measures for speech recognition.'' in \emph{Proc.
  Eurospeech}, 1997.

\end{thebibliography}

\end{document}